\documentclass[10pt,twocolumn,letterpaper]{article}

\usepackage{cvpr}
\usepackage{times}
\usepackage{epsfig}
\usepackage{graphicx}
\usepackage{amsmath}
\usepackage{amssymb}

\usepackage[table,xcdraw]{xcolor}

\usepackage[pagebackref=true,breaklinks=true,letterpaper=true,colorlinks,bookmarks=false]{hyperref}

\cvprfinalcopy % *** Uncomment this line for the final submission

\def\cvprPaperID{3545} % *** Enter the CVPR Paper ID here

\ifcvprfinal\fi
\begin{document}

%%%%%%%%% TITLE
\title{On Attention Modules for Audio-Visual Synchronization}

\author{Naji Khosravan\thanks{Work done during an internship at Netflix.}\\
\small{Center for Research in Computer Vision (CRCV)} \\
\small{University of Central Florida, Orlando, FL, USA}\\
{\tt\small najikh@cs.ucf.edu}
\and
Shervin Ardeshir\\
\small{Netflix, Inc}\\
\small{Los Gatos, CA, USA}\\
{\tt\small sardeshirbehrostaghi@netflix.com}
\and
Rohit Puri\\
\small{Netflix, Inc}\\
\small{Los Gatos, CA, USA}\\
{\tt\small rpuri@netflix.com}
}

\maketitle

%\thispagestyle{empty}

%%%%%%%%% ABSTRACT
\begin{abstract}
With the development of media and networking technologies, multimedia applications ranging from feature presentation in a cinema setting to video on demand to interactive video conferencing are in great demand. Good synchronization between audio and video modalities is a key factor towards defining the quality of a multimedia presentation. The audio and visual signals of a multimedia presentation are commonly managed by independent workflows - they are often separately authored, processed, stored and even delivered to the playback system. This opens up the possibility of temporal misalignment between the two modalities - such a tendency is often more pronounced in the case of produced content (such as movies) \footnote{There are several steps involved in going from live action camera shots to the finished presentation including video editing, special effects, audio mixing and dubbing.}. 

To judge whether audio and video signals of a multimedia presentation are synchronized, we as humans often pay close attention to discriminative spatio-temporal blocks of the video (e.g. synchronizing the lip movement with the utterance of words, or the sound of a bouncing ball at the moment it hits the ground). At the same time, we ignore large portions of the video in which no discriminative sounds exist (e.g. background music playing in a movie). Inspired by this observation, we study leveraging attention modules for automatically detecting audio-visual synchronization. We propose neural network based attention modules, capable of weighting different portions (spatio-temporal blocks) of the video based on their respective discriminative power. Our experiments indicate that incorporating attention modules yields state-of-the-art results for the audio-visual synchronization classification problem.
\end{abstract}

%%%%%%%%% BODY TEXT
\section{Introduction}
\begin{figure}[t!]
    \centering
    \includegraphics[width=1.0\columnwidth]{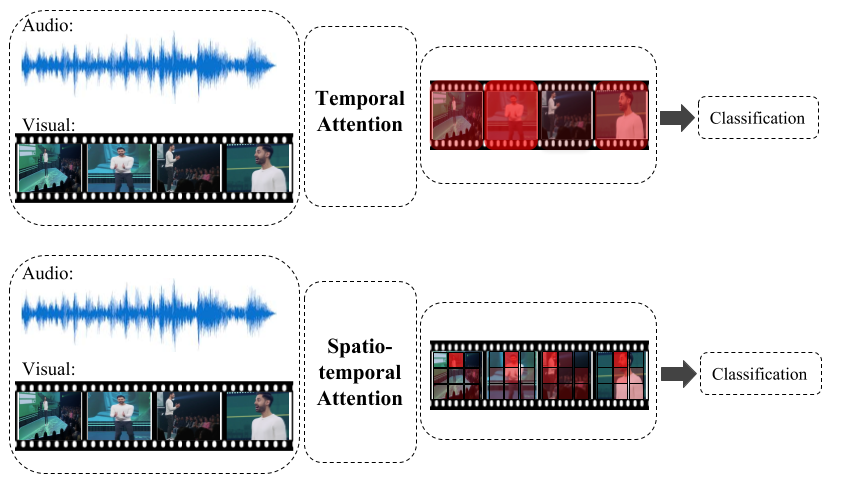}
    \caption{A high-level overview of the proposed approach. Top shows the overall framework for taking into account the temporal attention, where each temporal block is evaluated and its corresponding feature is weighted. Bottom shows the spatio-temporal attention module, a generalization of the temporal attention mechanism. Here the goal is to weight each spatio-temporal block of the video and make the decision based on the weighted features.}
    \label{fig:intro}
\end{figure}

Rapid progress in the domain of media, networking and playback technologies has resulted in a proliferation of multimedia applications. These range from feature presentation in a cinema setting to video on demand services over the Internet to real-time applications such as live video streaming applications as well as interactive video conferencing. Good synchronization between audio and video modalities is a significant factor towards defining the quality of a multimedia presentation. As an example, ``audio video not in sync'' is a major customer issue that video over IP providers have to deal with especially in a patchy connection bandwidth scenario. 

In a typical multimedia presentation, audio and video signals are managed by independent workflows. This is especially true in the case of highly produced premium content such as movies. In this case, the effort involved in going from raw audio and video (that were captured at the time of live action camera shots) to the finished work, is even managed via separate business processes. Examples of some of the steps that are involved here include video editing (deciding which portions of the raw footage will make it to the finished work), digital special effects (that are independently created for both video and audio), as well as sound mastering and mixing. Furthermore, a title intended for a global release often has several audio tracks corresponding to different international languages. Dubbed audio tracks are often created after the video and the primary language audio have been finalized.  All of this opens up the possibility of temporal mis-alignment between audio and video. The high costs involved in making a movie as well as the size of the intended audience warrant a high multimedia quality, necessitating tight synchronization between the audio and video media. 

As humans, we are highly capable of determining whether audio and visual signals of a multimedia presentation are synchronized. We seamlessly realize when and where to pay attention to in a video, in order to be able to judge whether there is a misalignment between audio and video.  We often pay close attention to spatio-temporal details such as lip movement of humans while watching a movie, and make our decision solely based on such cues. At the same time, we ignore large spatio-temporal portions of video, when we cannot find the source of the sound in the visual content (e.g., background music in a movie scene). Our ability to attend to the important details while ignoring the rest enables us to have a better judgment. 

A movie scene in which the visual content shows a person putting a coffee cup on the table, could contain many unrelated audio sources such as background noise or music. In principle, correctly identifying the sound of the coffee mug being placed on the table, and relating that to the visual content by focusing on the spatial location of the mug while ignoring the rest of the video, could lead to a very accurate audio-visual alignment prediction model. Correctly identifying when and where a bouncing basketball is hitting the ground could be another such example. In both these examples, the audio-visual phenomenon exhibits a high-degree of temporal and spatial localization. There are other scenarios such as a dialog or speech oriented multimedia presentation, where there are a lot of informative moments, that we can use to identify the synchronization between the utterance of each single word, and the lip movement of the speaker. In this effort, we aim to study the possibility of learning an attention model that is able to address all of the scenarios outline above. To do so, we use a convolutional neural network (CNN) based architecture that is capable of identifying the important portions of a multimedia presentation, and use them to determine the synchronization between the audio and visual signals. We study whether introducing attention modules would help the network emphasize on corresponding parts of the input data in order to make a better decision. 

To conduct this study, we define the problem of audio-video synchronization as a binary classification problem. Given a video, a fully convolutional network is trained to be able to decide whether the audio and visual modalities of the video are synchronized with each other or not. In order to train the network for this task, we expose the network to synchronized and non-synchronized audio-video streams alongside their binary labels during training time. We evaluate two different attention modules taking into account temporal only and spatial plus temporal dimensions respectively. As mentioned above, there are huge variations in multimedia data in terms of discriminativity for synchronization. While there are some scenes such as ocean waves that do not exhibit much spatial or temporal localization, there are others such as a bouncing ball that could exhibit temporal localization and others such as speech that could exhibit good spatial as well as temporal localization. We employ a soft-attention module, weighting different blocks of the video without enforcing the network to make hard decisions on each individual blocks.  

In order to take temporal attention into account, we divide each video into temporal blocks. From each temporal block of the video, we extract joint audio-visual features. We then compute a global feature pooled across the spatial and temporal domain within the block. Features from the various blocks are passed through a temporal weighting (attention) module, where the network assigns a confidence score to that temporal block of the video. The confidences for different temporal blocks of the video are normalized using a softmax function, enforcing the notion of probability across all the weights. The final decision about the alignment of the video is then made based on the weighted mean of the features extracted from different temporal blocks. Our experiments suggest that incorporating this attention module leads to higher classification accuracy and faster convergence rate compared to the baseline model (without any attention module).

We also study the effect of incorporating a more general spatio-temporal attention module on the classification accuracy. In this setup, the network is seeking to distinguish between not only different temporal blocks of the video, but also different spatial blocks of the visual content. To do so, similar to the first approach, we extract joint spatio-temporal features from each temporal block. Here however, instead of performing global average pooling, spatial features within each temporal block are directly fed into the weighting module, calculating a confidence score for each spatial block within each temporal block. Similar to the previous approach, a softmax function is applied across all the spatial and temporal features, and the final feature is computed as their weighted mean. 

In this paper we propose an attention based framework, trained in a self-supervised manner, for the audio-visual synchronization problem. The proposed attention modules learn to determine what to attend to in order to decide about the audio-visual synchrony of the video in the wild. We evaluate the performance of each of the two approaches on publicly available data in terms of classification accuracy. We observe that taking into account temporal and spatio-temporal attention leads to improvement in classification accuracy. We also evaluate the performance of the attention modules qualitatively, verifying that the attention modules are correctly selecting discriminative parts of the video.

% \begin{figure}[ht]
%     \centering
%     \includegraphics[width=1.0\columnwidth]{latex/Figures/sync_intro.pdf}
%     \caption{Caption}
%     \label{fig:intro}
% \end{figure}

\section{Related Work}
 To the best of our knowledge, this is the first attempt in using attention models for the audio-visual synchronization problem. Our approach could be classified as a self-supervised learning approach for utilizing attention models in the audio-visual synchronization problem. In the following, we go over recent works in the area of self-supervised learning, audio-visual synchronization, and attention models.
\subsection{Self-supervised Learning}
Data labeling and annotation in ``big data'' era can be an onerous task. It is common knowledge that performance of the deep neural network based learning approaches improves drastically with task relevant training data, however availability of good quality training data is scarce and it can be a very expensive and time consuming proposition to create the same. Fortunately, for some problem domains, self-supervised learning techniques can be employed to effectively create large amounts of ``labeled'' training data from input data that has no annotations.

A general framework for feature representation learning provided in \cite{noroozi2016unsupervised}. The impact of this framework has been shown on a variety of problems like object recognition, detection and segmentation. The main idea of \cite{noroozi2016unsupervised} is to make a Jigsaw puzzle on any given image and learn features that can solve the puzzles.  

Authors in \cite{liu2018leveraging} use self-supervised learning to produce relative labeled data for crowd counting problem. They crop each image and assume that the number of people in the cropped image is less than the original image. Finally, they train their network with a triplet loss and then fine-tune it with a limited number of supervised data.

Similarly, ranking loss as a self-supervision clue is discussed in \cite{liu2017rankiqa} to solve image quality assessment. The main idea of \cite{liu2017rankiqa} is to apply noise and blurring filters to natural images and solve a ranking problem between the original images and the noisy ones.

In this paper, we use self-supervised learning as a tool to train our deep neural network. Our work is most similar to the idea of \cite{liu2017rankiqa}. We shift the audio signal corresponding to video frames, randomly, to create training samples. Assuming audio and visual content are in sync in the original video (positive samples), we generate our negative samples by randomly shifting the audio signals for each positive sample.
 
\subsection{Audio-visual Synchronization}

As was discussed earlier audio-visual synchronization is a key factor for determining the quality of a multimedia presentation. It also has many real-world applications such as multimedia over network communications \cite{teng2000synchronization,lankford1995audio}, lip-reading \cite{chung2017lip}, and so on.

A similar work on the audio-visual synchronization is represented in~\cite{marcheret2015detecting}, in which, a simple Deep Neural Network has been used to detect the audio and video delays. A simple concatenation between audio and visual features has performed and the videos are recorded in a studio environment. Another related work to ours is that of lip synchronization \cite{chung2016out}. As opposed to our method, in \cite{chung2016out} authors used a two stream network with a contrastive loss. Lip synchronization is an audio-video synchronization problem limited to the domain of monologue face scenes. Faces are tracked and the mouths is cropped and then fed to the network. In this paper, we use more realistic data.

Synchronization is also an important first step for many other applications such as Lip-reading \cite{chung2017lip}. In this case, given an audio and a close-up video shot, the task is to predict the words of the speech happening in the video. 

We argue that in this paper we solve a more general problem dealing with unconditioned input data from real world examples uploaded to YouTube. A video can have many different types of scenes as well as abrupt view changes, different types of audio noises, etc. Our proposed attention modules decide where to attend in the video eliminating the need for further restrictions on the input.

The most similar work to this paper is~\cite{owens2018audio} where a self supervised method is used to train a network to classify sync and un-sync samples. A 3D convolutional network with early fusion of the two modalities is used. This work, however, does not have any notation of attention. On the other hand, our work is more focused on studying the contribution of temporal and spatio-temporal attentions in a similar problem setup. To the best of our knowledge, ours is the first work to use attention mechanisms for the audio-visual synchronization problem.

%Video and text data alignment is another problem that is related to the audio-visual synchronization problem. In both cases, there are two information modalities and also in both cases there is a temporal order in each stream of data (like frames of a video). Authors in \cite{Dogan18neumatch} solve the video-text alignment by using a Recurrent Neural Network (RNN) to decide at each time step if the textual input should be matched with the current shot or skip the current shot, merge the consecutive shots, or merge the consecutive texts. Similar to \cite{chung2016out,owens2018audio}, authors in \cite{Dogan18neumatch} do not use visual attention in their network. Also, the natural audio signals are much more noisy compared to text video descriptions, thus the audio-visual synchronization problem needs a more sophisticated solution.

%Furthermore, it is shown in \cite{korbar2018co} that synchronization of visual and audio streams needs a high level understanding of both domains. Authors in \cite{korbar2018co} show that the features extracted from a self-supervised audio-visual synchronization system, improves the results of other tasks like action recognition.

\subsection{Attention Models}
Attention mechanisms have been one of the popular set of techniques applied to various AI applications, namely, Question Answering (QA) \cite{sukhbaatar2015end}, Visual Question Answering (VQA) \cite{mazaheri2017video,yang2016stacked}, Visual Captioning \cite{xu2015show}, Machine Translation(MT) \cite{bahdanau2015mtbyalign, luong2015effective, vaswani2017attention}, Action Recognition  \cite{sharma2015action,zang2018attention}, as well as Robotics \cite{abolghasemi2018pay}.

Authors in \cite{xu2015show} show that to generate a description sentence for a given image, each word  relates to a spatial region of the image. Several attention maps were generated as they build the sentence. In machine translation \cite{luong2015effective}, attentions can help the translator network to align the source sentence phrases or words to a word in the target language sentence. Our task differs from  QA/VQA problems \cite{mazaheri2017video,yang2016stacked,sukhbaatar2015end}, in which given a question in form of a sentence, correlation between visual/textual features from different pieces of a video/image/text is computed as attention scores.

There are two main types of attention-based approaches: soft attention and hard attention. Hard attention methods \cite{weston2014memory,xu2015show} outputs are based on sampling decision for each point as being attended or not (binary). These methods mostly need ground truth of attended points, annotated by humans. On the other hand, soft attention methods \cite{vaswani2017attention,mazaheri2017video,sukhbaatar2015end} are able to capture the importance of the points in the given data, in terms of probability scores, to reach a given objective. Data points can be frames or shots in a video, audio segments, words in a sentence, or spatial regions in an image. Also, soft attention methods use differentiable loss functions and mathematical operations while hard attention approaches may not have continuous gradients. 

The attention mechanism used in this paper falls in the soft attention category since we provide a differentiable loss function, and our attention network produces a probability map over spatio-temporal or temporal video segments. The most similar work to our framework is \cite{zang2018attention}, in which authors use an attention network to select best shots of a video for action recognition based on separate streams of information. Similarly, in this paper, we apply attention modules on joint representations of audio-visual data in different blocks of a video. In contrast to \cite{zang2018attention}, in which all the streams of data represent visual information, in this work, we deal with two different data modalities.

We argue that only certain portions of the input data (temporal/spatio-temporal segments of a video) are useful for deciding about synchronization in that we can measure the alignment of audio and video solely based on them. In contrast to \cite{mazaheri2017video,yang2016stacked,sukhbaatar2015end}, a strong/weak correlation does not mean a strong attention in our work. For example, a view of ocean with the background sound of waves may always have a high correlation in feature space, but doesn't mean that it is a good shot to decide if audio and video are in sync. However, a close shot of a monologue speaker with a lot of background noises can be a very good shot to detect the synchronization.
\section{Framework}
%https://docs.google.com/drawings/d/1FuzGQQjnOpV9Q3TiVp4WqJG9Ik-eDk44QtnCnuEfb2M/edit?usp=sharing

\begin{figure*}[ht]
    \centering
    \includegraphics[width=\textwidth]{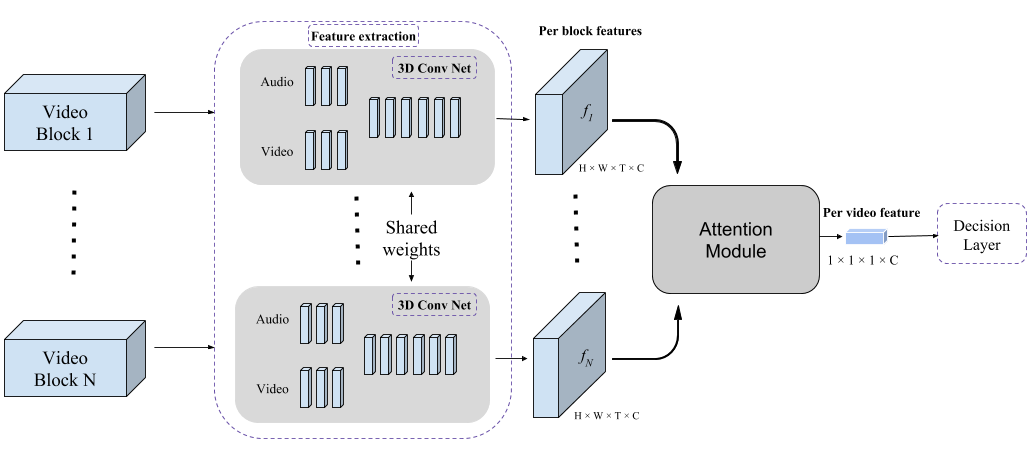}
    \caption{Architecture of the proposed approach. The video is split into several 25 frame temporal blocks. Each temporal block is passed through a 3D convolutional neural network, extracting spatio-temporal features from each temporal block. The features are the input to the attention module, where they are evaluated in terms of discriminative power and combined into one video-level global feature. The decision is finally made based on the global feature.}
    \label{fig:Architecture_1}
\end{figure*}

Our proposed neural network architecture involves three main steps. The first step is a feature extraction step. Here, we split the input video into several blocks and extract joint audio-visual features from each block. In the second step, we calculate temporal or spatio-temporal attention, evaluating the importance of different (temporal or spatio-temporal) parts of the video. Finally, we combine features extracted from different parts of the video into a per video global feature based on a weighted average of all the features. In the following, we provide details on the data representation, the two architectures used for temporal and spatio-temporal attention, and the training and testing procedures. 

\subsection{Joint Representation}
\label{sec:joint_representation}
The backbone of our architecture, is that of \cite{owens2018audio}. As shown in Figure \ref{fig:Architecture_1}, we divide the input video into $N$ non-overlapping temporal blocks of length $25$ frames (approximately $0.8$ secsonds given the frame rate of $29.97$ for our videos), namely $B_1,B_2,...,B_N$. We extract a joint audio-visual feature from each temporal block resulting in a $H \times W \times T \times C$ tensor $f_i$ for block $B_i$. $f_i$ is obtained by applying the convolutional network introduced in \cite{owens2018audio}, where visual and audio features are extracted separately in the initial layers of the network and later concatenated across channels. The visual features result in a $H \times W \times T \times C_v$ feature and the audio feature results in a $T \times C_a$ feature. The audio feature is replicated $H \times W$ times and concatenated with the visual feature across channels, resulting in a $H \times W \times T \times (C_v+C_a)$ dimensional tensor where $C = C_v+C_a$. The network is followed by $5$ convolution layers applied to the concatenated features, combining the two modalities and resulting in a joint representation. The joint representation is the input to the attention modules. We describe the details of applying temporal and spatiotemporal attention modules in the following sections.   

\subsection{Attention Modules}
\label{sec:attention_modules}
Our attention modules consist of two layers of $1 \times 1 \times 1$ convolutions applied to the joint audio-visual features, resulting in a scalar confidence value per block (temporal or spatio-temporal). The confidences are then passed through a softmax function to obtain a weight for each of these blocks. The weights are used to obtain a weighted mean of all the features of the video. The weighted mean is passed to the decision layer (as depicted in Figure \ref{fig:Architecture_1}). In other words, the attention modules evaluate each portion of the video (a temporal or spatio-temporal block) in terms of its importance and therefore, its contribution to the final decision. In the following, we will go over a more detailed description of the two attention modules studied in this work.

\subsubsection{Temporal Attention}
\label{sec:temporal_attention}
\begin{figure}[ht]
    \centering
    \includegraphics[width=.5\textwidth]{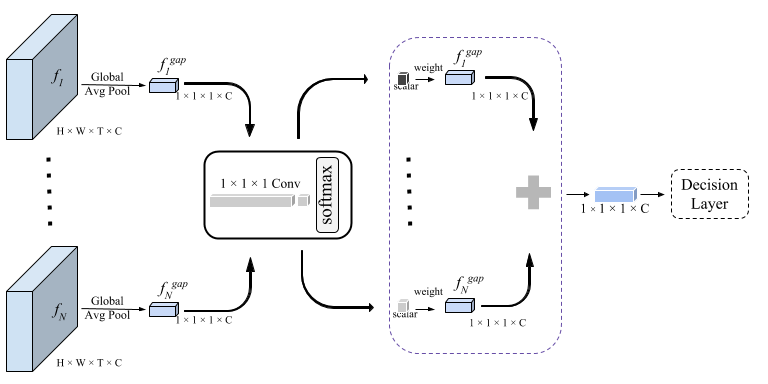}
    \caption{The temporal attention module: $1 \times 1 \times 1$ convolutions are used to obtain a confidence score for each single temporal block. All the confidences are passed through a softmax function and the resulting weights are applied to the temporal features ($1 \times 1 \times 1 \times C$). The weighted average of all features (a $C$ dimensional vector) is then passed through the decision layer.}
    \label{fig:Architecture_2}
\end{figure}

As explained in Section \ref{sec:joint_representation}, a video results in a set of features $f_1,f_2,...,f_N$. For the temporal attention module, we apply global average pooling to each $H \times W \times T \times C$ dimensional feature $f_i$ across spatial and temporal dimensions, resulting in a $1 \times 1 \times 1 \times C$ dimensional feature $f^{gap}_i$. Therefore, representing each block of the video using a single global feature vector $f^{gap}_i$. We apply $1 \times 1 \times 1$ convolution layers on the global average pooled features, resulting in a single scalar confidence value $c_i$ for each temporal block $B_i$. The confidence value $c_i$ is intuitively capturing the absolute importance of that specific temporal block. Applying a softmax normalization function over all the confidence values of different time-blocks of the same video, we obtain a weight $w_i$ for each feature $f^{gap}_i$. The normalization is performed to enforce the notion of probability and keep the norms of the output fearures in the same range as each individual global feature. The weighted mean of the features $\Sigma_{i} w_i f^{gap}_i$ is passed to the decision layer (see figure \ref{fig:Architecture_2}).  

\subsubsection{Spatio-temporal Attention}
\label{sec:spatio_temporal_attention}
\begin{figure}[ht]
    \centering
    \includegraphics[width=.5\textwidth]{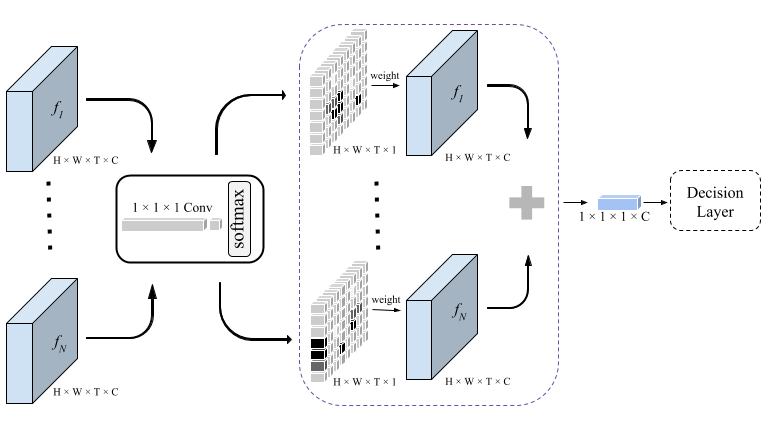}
    \caption{The spatio-temporal attention module: $1 \times 1 \times 1$ convolutions are used to obtain a confidence score for each single spatio-temporal block (each spatial block within each temporal block). All the confidences are passed through a softmax function and the resulting weights are applied to the spatio-temporal features ($1 \times 1 \times 1 \times C$). The weighted average of all features (a $C$ dimensional vector) is then passed through the decision layer.}
    \label{fig:Architecture_3}
\end{figure}

For the spatio-temporal attention module, we apply the $1 \times 1 \times 1$ convolution layers directly on the $H \times W \times T \times C$ dimensional features, resulting in a set of confidence values $c_{H \times W \times T}$ for each block. We then enforce the notion of probability across all the confidence values of all the blocks ($H \times W \times T$ scalar values). The decision is made based on the weighted average on the spatio-temporal features $\Sigma^N_{n=1}\Sigma^T_{i=1}\Sigma^H_{j=1}\Sigma^W_{k=1}w_{nijk}f^{ijk}_n$, where $f^{ijk}_n$ is a feature vector extracted from a single spatial block $i,j,k$ at temporal block $n$ (see figure \ref{fig:Architecture_3}). 

\subsection{Baseline}
\label{sec:baseline}
In order to evaluate the effect of our temporal and spatio-temporal attention modules, we compare their performance with the performance of a uniform weighting baseline. As the attention modules simply calculate weights for the features, and the decision is made based on the weighted average of those features, as a baseline, we simply feed the average of the input features directly into the decision layer. In other words, we evaluate the effect of bypassing the weighting step.

\subsection{Implementation Details}
\label{sec:implementation_details}
Here we explain the implementation details of the feature extraction step, the two attention modules (temporal and spatio-temporal), and the decision layer used in our work.

\subsubsection{Input and Feature Extraction Step}
The feature extraction network consists of a 3D convolutional neural network with an early-fused design. In order to have a fair comparison with the baseline network, we use a similar setup as \cite{owens2018audio}, meaning that the input video is resized to $256 \times 256$ and center cropped to $224 \times 224$. Video lengths are 4.2 second and the lenbgth of each temporal block is 25 frames. All the videos have frame rate of 29.97 Hz. The amount of shift on the audio signal for the negative examples is a randomly generated value from 2 to 5.8 seconds.  

\subsubsection{Temporal Attention Module}
The temporal attention module consist of two $1 \times 1 \times 1$ convolution layers with $128$ and $1$ channels respectively, with relu activation and dropout ratio of $0.5$. The convolutions are applied to the globally average pooled feature extracted from the input video. 

\subsubsection{Spatio-temporal Attention Module}
Similar to the temporal attention module, two $1 \times 1 \times 1$ convolution layers here with $16$ and $1$ channels, respectively. The dropout ratio is $0.4$. Unlike the temporal attention module, the spatio-temporal attention module is applied to the spatial features directly (without global average pooling).

\subsubsection{Decision Layer}
For the sake of a fair comparison, we use the same architecture for the decision layers of all three networks (base network with uniform weighting, temporal and spatio-temporal attention modules). It consist of $2$ layers of fully connected (or equivalently convolution with kernel size of $1$) of $512$ and $2$ dimensions for the binary classification. 
The output of the attention modules are passed through a softmax function, and used to obtain the weighted mean of the features, which is the input to the decision layer. 

\subsubsection{Training}
For all three networks (i.e, base network with uniform weighting, temporal weighting, and spatio-temporal weighting) we used the pre-trained weights of \cite{owens2018audio} for our feature extraction step, freezing the early layers' weights. We trained the decision layer and attention modules from scratch with batch size of $80$ and for $300$ epochs. Binary cross-entropy loss, Adam optimizer, and learning rate of $1e-3$ was used for all three networks.  

\section{Experiments}
In this section, we go over the dataset used for training and evaluating the performance of the proposed approach in Section \ref{sec:dataset}. We report the quantitative results in Section \ref{sec:quantitative} and go over some qualitative examples in Section \ref{sec:qualitative}.   

\subsection{Dataset}
\label{sec:dataset}
We evaluate the proposed approach on the publicly AudioSet \cite{gemmeke2017audio} dataset, which contains an ontology of 632 audio event categories. We train the temporal and spatio-temporal modules on 3000 examples of the speech subset of the dataset, and test the proposed approach on 7000 examples of the speech dataset, and 800 examples from the generic impact sound categories to further show the robustness of our method on sound classes such as breaking, hitting, bouncing etc. in which attention plays an important role. We used each video as a positive example, and a misaligned version of the video as a negative example.  
\subsection{Quantitative Evaluation}
\label{sec:quantitative}
We evaluate the performance of the proposed approaches in terms of binary classification accuracy. The classification accuracies are reported in Table \ref{tab:classification_accuracy}. The first row shows the performance of the baseline method, where no attention module is used. Comparing the first two rows of the table, we can observe the effect of using temporal attention in the classification accuracy. We can see that in the speech category, using temporal attention leads to $4.9\%$ improvement in classification accuracy. In the generic sound class, temporal attention yields a higher accuracy boost of $9.3\%$. We attribute the lower margin in the speech class to the fact that in speech videos, most of the temporal blocks of the video do contain discriminative features (lip movement) and therefore, the weights are generally more uniform (see Figure \ref{fig:qualitative_temporal_t}). The last row shows the performance of our network with the spatio-temporal attention module. In the speech class, incorporating spatio-temporal attention leads to $3.8\%$ compared to using temporal attention, and $8.7\%$ compared to not incorporating attention at all and $9.8\%$ improvement on generic sound class. Spatio-temporal attention has a lower margin of improvement over temporal attention in generic sound class compared to speech. This lower margin could be associated to the fact that speech videos tend to be more spatially localized (towards the face of the speaker).

\begin{table}[h]
\begin{tabular}{|l|l|l|}
\hline
\rowcolor[HTML]{C0C0C0} 
 Method & Speech & Generic sound  \\ \hline
 Baseline network \cite{owens2018audio} & 0.716 & 0.658  \\ \hline
 Temporal attention & 0.765 & 0.751 \\ \hline
 Spatio-temporal attention & \textbf{0.803} & \textbf{0.756} \\ \hline
\end{tabular}
\caption{Classification accuracy: Left column contains the methods being evaluated in terms of binary classification accuracy. Middle column contains the performance of the evaluated approaches on the Speech category of the AudioSet \cite{gemmeke2017audio} dataset. Right column contains the performance on the Generic sound category. Comparing the performance of the Baseline network with temporal attention, shows the effect of the temporal attention module. The effect of using spatio-temporal attention could be observed by comparing the accuracies reported in the last two rows.}
\label{tab:classification_accuracy}
\end{table}

\begin{figure}
    \centering
    \includegraphics[width=1.0\columnwidth]{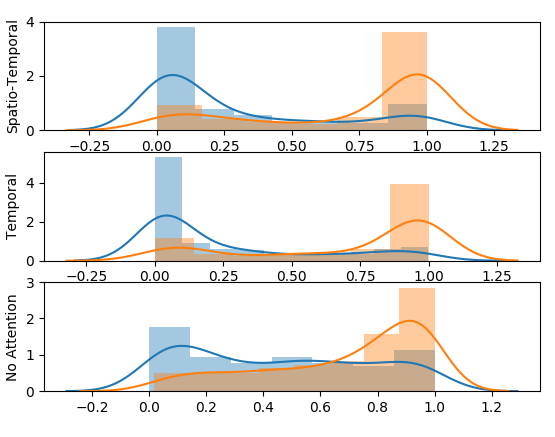}
    \caption{Comparing the distribution of scores predicted by the proposed approach, and comparing it to the baseline. The distribution of the scores obtained from negative examples (non-synced videos) are shown in blue, and the distribution of the scores for the aligned videos are shown in orange. It can be observed that using attention modules causes more successful separation of positive and negative examples by the network. }
    \label{fig:Prob_dist}
\end{figure}

To further illustrate the effect of attention modules, we plot and compare the distributions of output scores from our classification network in Figure \ref{fig:Prob_dist}. As can be seen attention modules help for a better separation of the two classes of sync and un-sync data distributions.

% \begin{figure}
%     \centering
%     \includegraphics[width=1.0\columnwidth]{latex/Figures/weights_dist.png}
%     \caption{The distribution of the weights.}
%     \label{fig:intro}
% \end{figure}
\subsection{Qualitative Evaluation}
\label{sec:qualitative}
\begin{figure}[h!]
    \centering
    \includegraphics[width=1.0\columnwidth]{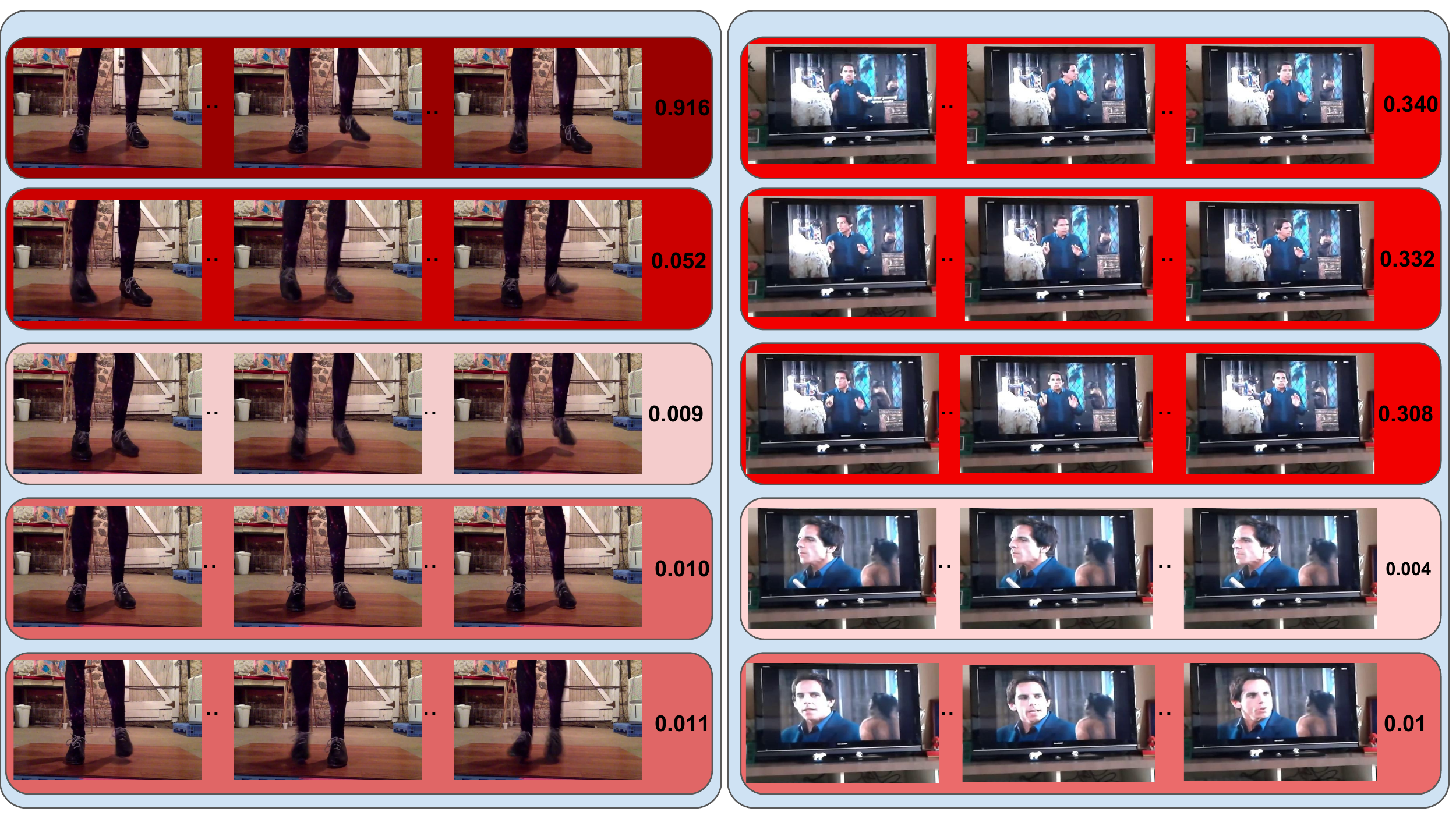}
    \caption{Qualitative examples from the temporal attention module: Each row shows a temporal block of a video, highlighted with its corresponding attention weight (color-coded).}
    \label{fig:qualitative_temporal_t}
\end{figure}

Here we visualize some examples of the weights estimated by the network. We expect the informative parts of the video to lead to higher values. Two examples of the temporal attention weights are shown in Figure \ref{fig:qualitative_temporal_t} (one from each class of dataset). In each example, each row contains one of the temporal blocks. We also show the score for each temporal block. As it can be observed, in the example on the left, a high weight has been assigned to the informative moment of the shoe tapping the ground. In the example on the right, the moments when the words are uttered by the actor are selected as the most informative parts. 

\begin{figure}[h!]
    \centering
    \includegraphics[width=1.0\columnwidth]{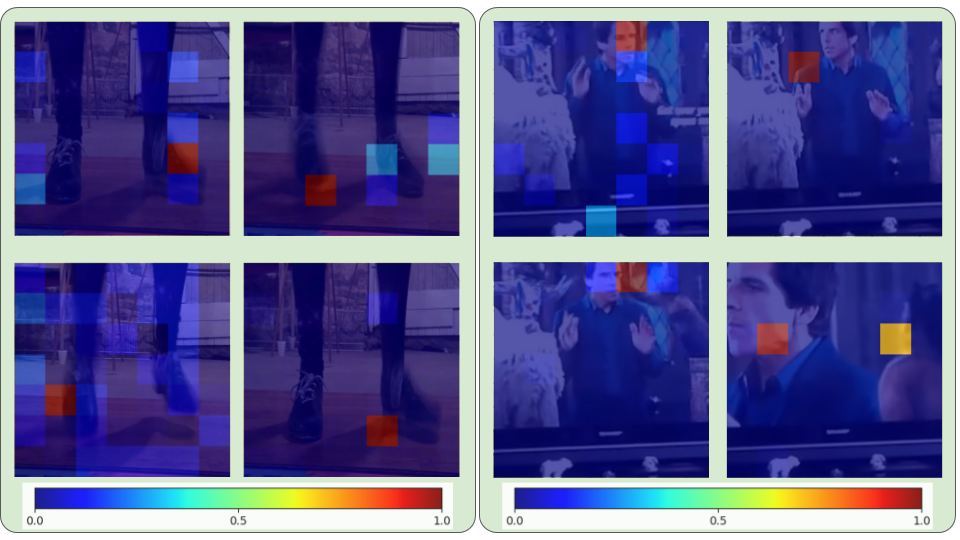}
    \caption{Qualitative examples from the spatio-temporal attention module. We picked the same examples for which we have visualized the temporal attention scores in Figure \ref{fig:qualitative_temporal_t}. It can be seen that the location of the shoe tapping on the floor and the face of the speaker are localized by the network.}
    \label{fig:qualitative_temporal_st}
\end{figure}

In Figure \ref{fig:qualitative_temporal_st}, we show the weights obtained from the spatio-temporal module on the same examples. It can be obsersved that the network correctly assigns higher values to more discriminative regions of the video (e.g. shoe tapping the floor, and the speakers face).

\section{Conclusion}
\label{sec:conclusion}
In this work we studied the effect of incorporating temporal and spatio-temporal attention modules in the problem of audio-visual synchronization. Our experiments suggest that a simple temporal attention module could lead to substantial gains in terms of classifying synchronized vs non-synchronized audio and visual content in a video. Also, a more general spatio-temporal attention module could even achieve higher accuracy as it is additionally capable of focusing on more discriminative spatial blocks of the video. Visualizing the weights generated by the temporal and spatio-temporal attention modules, we observe that the discriminative parts of the video are correctly given higher weights. 

To conclude, our experiments suggest that incorporating attention models in the audio-visual synchronization problem could lead to higher accuracy. Other variations of this approach, such as using different backbones for feature extraction, adopting different architectures such as recurrent models, could be potentially explored in the future. Furthermore, estimating the level of misalignment between the two modalities can be explored through a modification in the architecture. We believe that this work could be a first step in these directions.

{\small
\bibliographystyle{ieeetr}
\bibliography{egbib}
}

\end{document}